\documentclass[sigchi,authorversion]{acmart}

\usepackage{bookmark}
\usepackage{bm}
\usepackage{bbm}
\usepackage{amsmath}
\usepackage{amssymb}
\usepackage{booktabs}
\usepackage{mathtools}
\usepackage[labelfont=bf,textfont=bf]{subcaption}
\usepackage{enumitem}

\newcommand{\coollink}[1]{\href{https://#1}{\nolinkurl{#1}}}

\setcopyright{rightsretained}

\acmDOI{10.1145/3301275.3302288}

\acmISBN{978-1-4503-6272-6/19/03}

\acmConference[IUI '19]{24th International Conference on Intelligent User
Interfaces}{March 17--20, 2019}{Marina del Ray, CA, USA}
\acmBooktitle{24th International Conference on Intelligent User Interfaces (IUI '19),
March 17--20, 2019, Marina del Ray, CA, USA}
\acmYear{2019}
\copyrightyear{2019}

\acmPrice{15.00}

\begin{document}
\title{Piano Genie}

\author{Chris Donahue}
\authornote{Work performed while interning at Google AI.}
\orcid{1234-5678-9012}
\affiliation{%
  \institution{UC San Diego}
}
\email{cdonahue@ucsd.edu}

\author{Ian Simon}
\affiliation{%
  \institution{Google AI}
}
\email{iansimon@google.com}

\author{Sander Dieleman}
\affiliation{%
  \institution{DeepMind}
}
\email{sedielem@google.com}

\renewcommand{\shortauthors}{Donahue et al.}

\begin{abstract}
We present Piano Genie, an intelligent controller which allows non-musicians to improvise on the piano. With Piano Genie, a user performs on a simple interface with eight buttons, and their performance is decoded into the space of plausible piano music in real time. To learn a suitable mapping procedure for this problem, we train recurrent neural network autoencoders with discrete bottlenecks: an encoder learns an appropriate sequence of buttons corresponding to a piano piece, and a decoder learns to map this sequence back to the original piece. During performance, we substitute a user's input for the encoder output, and play the decoder's prediction each time the user presses a button. To improve the intuitiveness of Piano Genie's performance behavior, we impose musically meaningful constraints over the encoder's outputs.
\end{abstract}

%
%




\maketitle

\section{Introduction}

While most people have an innate sense of and appreciation for music, 
comparatively few are able to participate meaningfully in its creation. 
A non-musician could endeavor to achieve proficiency on an instrument, but the time and financial requirements may be prohibitive. 
Alternatively, a non-musician could operate a system which automatically generates complete songs at the push of a button, 
but this would remove any sense of ownership over the result. 
We seek to sidestep these obstacles by designing an intelligent interface which takes high-level specifications provided by a human and maps them to plausible musical performances.

The practice of ``air guitar'' offers hope that non-musicians can provide such specifications~\cite{godoy2005playing}
---
performers strum fictitious strings with rhythmical coherence and even move their hands up and down an imaginary fretboard in correspondence with \emph{melodic contours}, i.e.~rising and falling movement in the melody. 
This suggests a pair of attributes which may function as an effective communication protocol between non-musicians and generative music systems: 1)~rhythm, and 2)~melodic contours. 
In addition to air guitar, 
rhythm games such as \emph{Guitar Hero}~\cite{guitarhero} also make use of these two attributes. 
However, both experiences only allow for the imitation of experts and provide no mechanism for the \emph{creation} of music.

\begin{figure}
    \centering
    \includegraphics[width=0.99\linewidth]{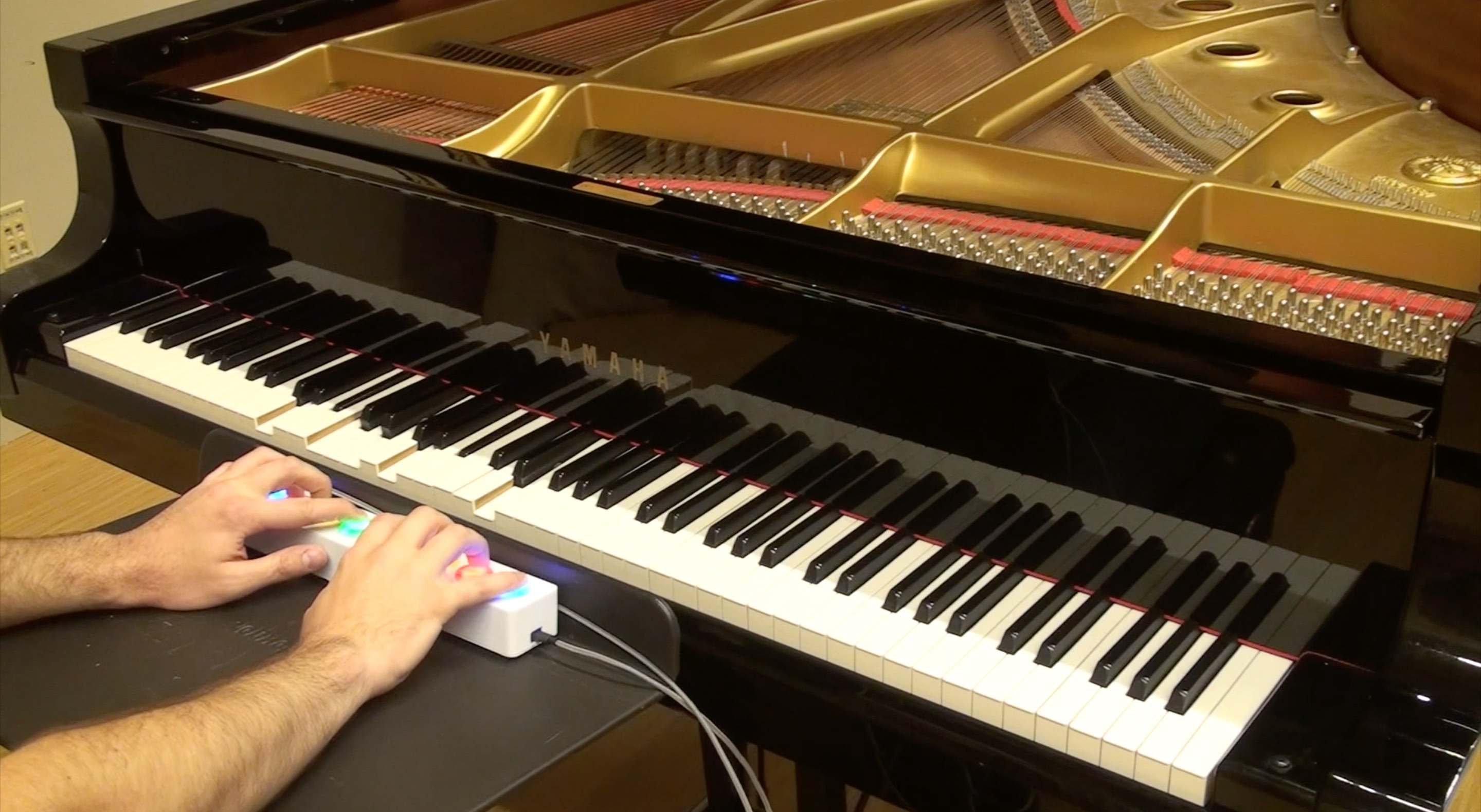}
    \caption{Using Piano Genie to improvise on a Disklavier (motorized piano) via MIDI. Video: \coollink{youtu.be/YRb0XAnUpIk}}
    \label{fig:interface}
\end{figure}

In this work, we present \emph{Piano Genie}, 
an intelligent controller allowing non-musicians to improvise on the piano while retaining ownership over the result (Figure~\ref{fig:interface}). 
In our web demo, a participant improvises on eight buttons, and their input is translated into a piano performance by a neural network running in the browser in real-time.\footnote{\textbf{Web Demo}: \coollink{piano-genie.glitch.me}, \textbf{Video}: \coollink{youtu.be/YRb0XAnUpIk}, \\
\textbf{Training Code}: \coollink{bit.ly/2Vv78Gx}, \textbf{Inference Code}: \coollink{bit.ly/2QolPrb}} 
Piano Genie has similar performance mechanics to those of a real piano: pressing a button will trigger a note that sounds until the button is released. 
Multiple buttons can be pressed simultaneously to achieve polyphony. 
The mapping between buttons and pitch is non-deterministic, 
but the performer can control the overall form by pressing higher buttons to play higher notes and lower buttons to player lower notes.

\begin{figure}[!ht]
    \centering
    \includegraphics[width=0.99\linewidth]{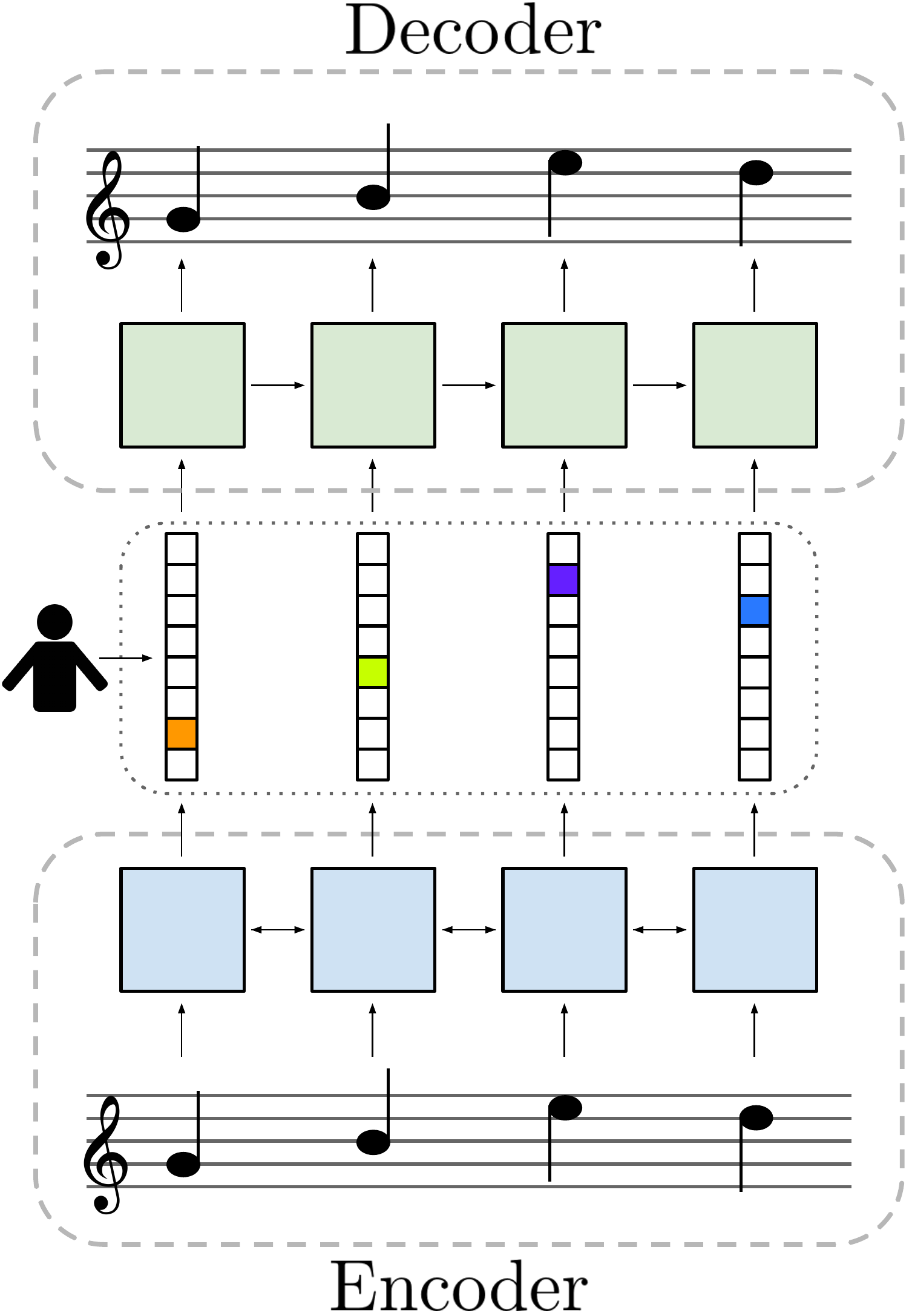}
    \caption{Piano Genie consists of a discrete sequential autoencoder. A bidirectional RNN encodes discrete piano sequences ($88$ keys) into smaller discrete latent variables ($8$ ``buttons''). The unidirectional decoder is trained to map the latents back to piano sequences. During inference, the encoder is replaced by a human improvising on buttons.}
    \label{fig:model}
\end{figure}

Because we lack examples of people performing on ``miniature pianos'', 
we adopt an unsupervised strategy for learning the mappings. 
Specifically, 
we use the \emph{autoencoder} setup, 
where an \emph{encoder} learns to map $88$-key piano sequences to $8$-button sequences, 
and a \emph{decoder} learns to map the button sequences back to piano music (Figure~\ref{fig:model}). 
The system is trained end-to-end to minimize reconstruction error.
At performance time, 
we replace the encoder's output with a user's button presses, 
evaluating the decoder in real time.

\section{Related Work}

Perception of melodic contour is a skill acquired in infancy~\cite{trehub1984infants}. 
This perception is important for musical memory and is somewhat invariant to transposition and changes in intervals~\cite{dowling1978scale,huron1996melodic}. 
The act of \emph{sound tracing}---moving one's hands in the air while listening to music---has been studied in music information retrieval~\cite{parsons1975directory,godoy2009body,nymoen2011analyzing,kelkar2017representation,olivier2018soundtracer}.
It has been suggested that the relationship between sound tracings and pitch is non-linear~\cite{eitan2014lower,kelkar2018evaluating}. 
Like Piano Genie, 
some systems use user-provided contours to \emph{compose} music~\cite{roy2014trap,kitahara2017jamsketch}, 
though these systems generate complete songs 
rather than allowing for real-time improvisation. 
An early game by Harmonix called \emph{The Axe}~\cite{theaxe} allowed users to improvise in real time by manipulating contours which indexed pre-programmed melodies.

There is extensive prior work~\cite{lee1992neural,bevilacqua2005mnm,fiebrink2009meta,gillian2011machine} on \emph{supervised} learning of mappings from different control modalities to musical gestures. 
These approaches require users to provide a training set of control gestures and associated labels. 
There has been less work on \emph{unsupervised} approaches, 
where gestures are automatically extracted from arbitrary performances. 
Scurto and Fiebrink~\cite{scurto2016grab} describe an approach to a ``grab-and-play'' paradigm, 
where gestures are extracted from a performance on an arbitrary control surface, and mapped to inputs for another. 
Our approach differs in that the controller is fixed and integrated into our training methodology, and we require no example performances on the controller.

\section{Methods}

We wish to learn a mapping from sequences ${\bm{y} \in [0, 8)^n}$, 
i.e.~amateur performances of $n$ presses on eight buttons, 
to sequences ${\bm{x} \in [0, 88)^n}$, 
i.e.~professional performances on an $88$-key piano. 
To preserve a one-to-one mapping between buttons pressed and notes played, 
we assume that both $\bm{y}$ and $\bm{x}$ are monophonic sequences.\footnote{This does not prevent our method from working on polyphonic piano music; 
we just consider each key press to be a separate event.}
Given that we lack examples of $\bm{y}$, we propose using the autoencoder framework on examples $\bm{x}$. 
Specifically, we learn a deterministic mapping ${\text{enc}(\bm{x}) : [0, 88)^n \mapsto [0, 8)^n}$, 
and a stochastic inverse mapping ${P_{\text{dec}}(\bm{x} | \text{enc}(\bm{x}))}$. 

\begin{figure}
\centering
\begin{subfigure}{.5\linewidth}
  \centering
  \includegraphics[width=.9\linewidth]{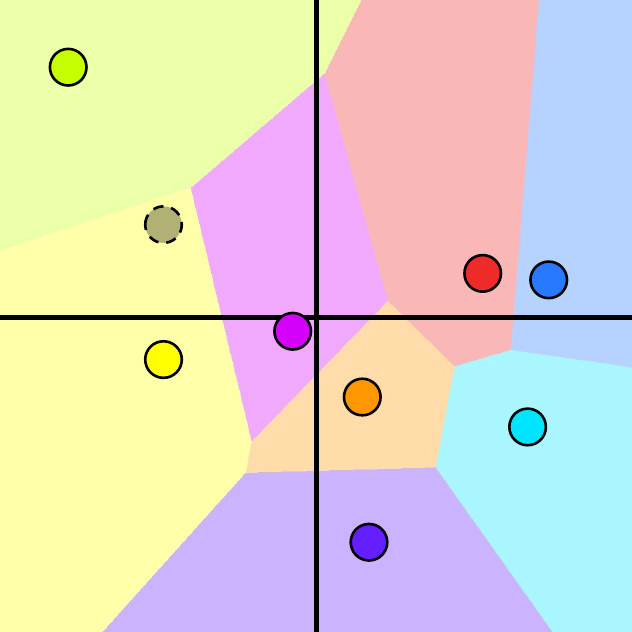}
  \caption{VQ-VAE~\cite{van2017neural} quantization}
  \label{fig:vqvae}
\end{subfigure}%
\begin{subfigure}{.5\linewidth}
  \centering
  \includegraphics[width=.9\linewidth]{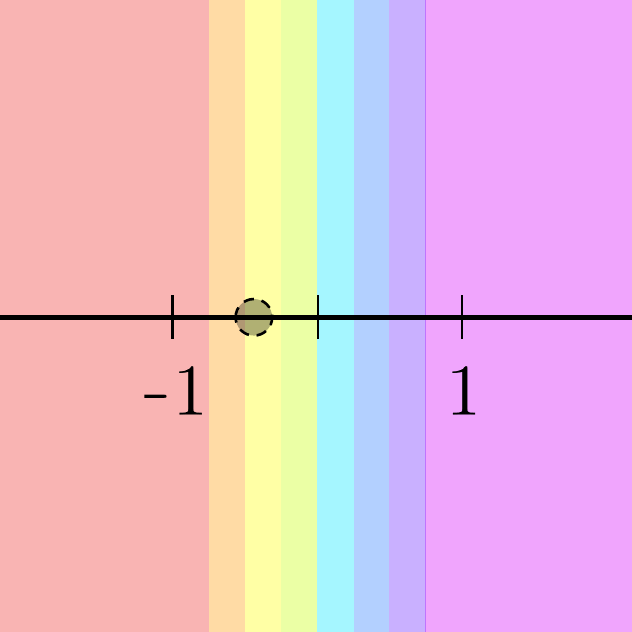}
  \caption{IQAE quantization}
  \label{fig:iqae}
\end{subfigure}
\caption{Comparison of the quantization scheme for two autoencoder strategies with discrete latent spaces. The VQ-VAE (left) learns the positions of $k$ centroids in a $d$-dimensional embedding space (in this figure $k=8,d=2$, in our experiments $k=8,d=4$). An encoder output vector (grey circle) is quantized to its nearest centroid (yellow circle) before decoding. Our IQAE strategy (right) quantizes a scalar encoder output (grey) to its nearest neighbor (yellow) among $k=8$ centroids evenly spaced between $-1$ and $1$.}
\label{fig:vqvae_vs_iqae}
\end{figure}

We use LSTM recurrent neural networks (RNNs)~\cite{hochreiter1997long} for both the encoder and the decoder. 
For each input piano note, the encoder outputs a real-valued scalar, 
forming a sequence $\text{enc}_{s}(\bm{x}) \in \mathbb{R}^n$. 
To discretize this into $\text{enc}(\bm{x})$
we quantize it to $k=8$ buckets equally spaced between $-1$ and $1$ (Figure~\ref{fig:iqae}), 
and use the straight-through estimator~\cite{bengio2013estimating} to bypass this non-differentiable operation in the backwards pass. 
We refer to this contribution as the integer-quantized autoencoder (IQAE); 
it is inspired by two papers from the image compression literature that also use autoencoders with discrete bottlenecks~\cite{balle2016end,theis2017lossy}. 
We train this system end-to-end to minimize: 
\begin{equation}
L = L_{\text{recons}} + L_{\text{margin}} + L_{\text{contour}}
\vspace{-3mm}
\end{equation}
\begin{align*}
L_{\text{recons}} &= -\Sigma\log P_{\text{dec}}(\bm{x} | \text{enc}(\bm{x})) \\
L_{\text{margin}} &= \Sigma\max(|\text{enc}_{s}(\bm{x})| - 1, 0)^2 \\
L_{\text{contour}} &= \Sigma\max(1-\Delta{}\bm{x}\Delta{}\text{enc}_{s}(\bm{x}), 0)^2
\end{align*}

Together, $L_\text{recons}$ and $L_\text{margin}$ constitute our proposed IQAE objective. 
The former term minimizes reconstruction loss of the decoder (as is typical of autoencoders). 
To agree with our discretization strategy, the latter term discourages the encoder from producing values outside of ${[-1, 1]}$. 
We also contribute a musically motivated regularization strategy
which gives the model an awareness of melodic contour. 
By comparing the finite differences (musical intervals in semitones) of the input $\Delta{}\bm{x}$ to the finite differences of the real-valued encoder output $\Delta{}\text{enc}_{s}(\bm{x})$, 
the $L_\text{contour}$ term encourages the encoder to produce ``button contours'' that match the shape of the input melodic contours. 

\section{Experiments and Analysis}

\begin{table}
\centering
\caption{Quantitative results comparing an RNN language model, the VQ-VAE ($k=8,d=4$), and our proposed IQAE model ($k=8$) with and without contour regularization. $\Delta{}T$ adds time shift features to model. PPL is perplexity: $e^{L_{\text{recons}}}$. CVR is contour violation ratio: the proportion of timesteps where the sign of the melodic interval $\neq$ that of the button interval. Gold is the mean squared error in button space between the encoder outputs for familiar melodies and manually-created gold standard button sequences for those melodies. Lower is better for all metrics.}
\begin{tabular}{lccc}
\toprule
Configuration  & PPL & CVR & Gold \\
\midrule
Language model               & $15.44$ &  &   \\
~~$+\Delta{}T$               & $11.13$ &  &   \\
VQ-VAE~\cite{van2017neural}  & $3.31$ & $.360$ & $9.69$  \\
~~$+\Delta{}T$               & $2.82$ & $.465$ & $9.15$   \\
IQAE                         & $3.60$ & $.371$ & $5.90$  \\
~~$+L_{\text{contour}}$      & $3.53$ & $.002$ & $1.70$ \\
~~$+L_{\text{contour}} + \Delta{}T$ & $3.16$ & $.004$ & $1.61$  \\
\bottomrule
\end{tabular}
\label{tab:results}
\end{table}

We train our model on the Piano-e-Competition data~\cite{pianoe}, 
which contains around $1400$ performances by skilled pianists. 
We flatten each polyphonic performance into a single sequence of notes ordered by start time, breaking ties by listing the notes of a chord in ascending pitch order.
We split the data into training, validation and testing subsets using an $8:1:1$ ratio. 
To keep the latency low at inference time, 
we use relatively small RNNs consisting of two layers with $128$ units each. 
We use a bidirectional RNN for the encoder, and a unidirectional RNN for the decoder since it will be evaluated in real time. 
Our training examples consist of $128$-note subsequences randomly transposed between $[-6, 6)$ semitones. 
We perform early stopping based on the reconstruction error on the validation set.

As a baseline, we consider an LSTM ``language model''---equivalent to the decoder portion of our IQAE without button inputs---trained to simply predict the next note given previous notes. 
This is a challenging sequence modeling task, among other reasons because the monophonic sequences will frequently jump between the left and the right hand. 
To allow the network to factor in timing into its predictions, 
we add in a $\Delta{}T$ feature to the input, 
representing the amount of time since the previous note quantized into $32$ buckets evenly spaced between $0$ and $1$ second. 
This language model baseline is not unlike our previous work on Performance RNN~\cite{performancernn2017}, 
though in that work our goal was to predict not only notes but also timing information and dynamics (here, timing is provided and dynamics are ignored).



We also compare to the VQ-VAE strategy~\cite{van2017neural}, 
an existing discrete autoencoder approach. 
The VQ-VAE strategy discretizes based on proximity to learned centroids within an embedding vector space (Figure~\ref{fig:vqvae}) as opposed to the fixed scalar centroids in our IQAE (Figure~\ref{fig:iqae}). 
Accordingly, it is not possible to apply the same contour regularization strategy to the VQ-VAE, and the meaning of the mapping between the buttons and the resultant notes is less interpretable.

\subsection{Analysis}
To evaluate our models, we calculate two metrics on the test set: 
1)~the perplexity (PPL) of the model $e^{L_{\text{recons}}}$, 
and 2)~the ratio of contour violations (CVR), i.e.~the proportion of timesteps where the sign of the button interval disagrees with the sign of the note interval. 
We also manually create ``gold standard'' button sequences for eight familiar melodies (e.g. \emph{Fr\`ere Jacques}), and measure the mean squared error in button space between these gold standard button sequences and the output of the encoder for those melodies (Gold).
We report these metrics for all models in Table~\ref{tab:results}.

\begin{figure}
    \centering
    \includegraphics[width=0.98\linewidth]{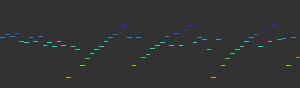}\\
    \vspace{1mm}
    \includegraphics[width=0.98\linewidth]{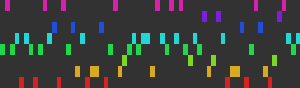}\\
    \vspace{1mm}
    \includegraphics[width=0.98\linewidth]{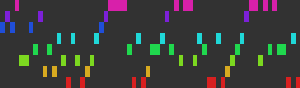}
    \caption{Qualitative comparison of the $8$-button encodings for a given melody (top) by the VQ-VAE (middle) and our IQAE with $L_{\text{contour}}$ (bottom). Horizontal is note index. The encoding learned by the IQAE echoes the contour of the musical input.}
    \label{fig:qualitative}
\end{figure}

As expected, 
all of the autoencoder models outperformed the language model in terms of reconstruction perplexity.
The VQ-VAE models achieved better reconstruction costs than their IQAE counterparts, but produced non-intuitive button sequences as measured by comparison to gold standards. 
In Figure~\ref{fig:qualitative}, we show a qualitative comparison between the button sequences learned for a particular input by the VQ-VAE and our IQAE with contour regularization. 
The sequences learned by our contour-regularized IQAE model are visually more similar to the input.

Interestingly, the IQAE model regularized with the $L_\text{contour}$ penalty had better reconstruction than its unregularized counterpart. 
It is possible that the contour penalty is making the decoder's job easier by limiting the space of mappings that the encoder can learn. 
The $L_\text{contour}$ penalty was effective at aligning the button contours with melodic contours; the encoder violates the melodic contour at less than $1\%$ of timesteps. 
The $\Delta{}T$ features improved reconstruction for all models.

\section{User Study}

\begin{figure}
    \centering
    \includegraphics[width=0.99\linewidth]{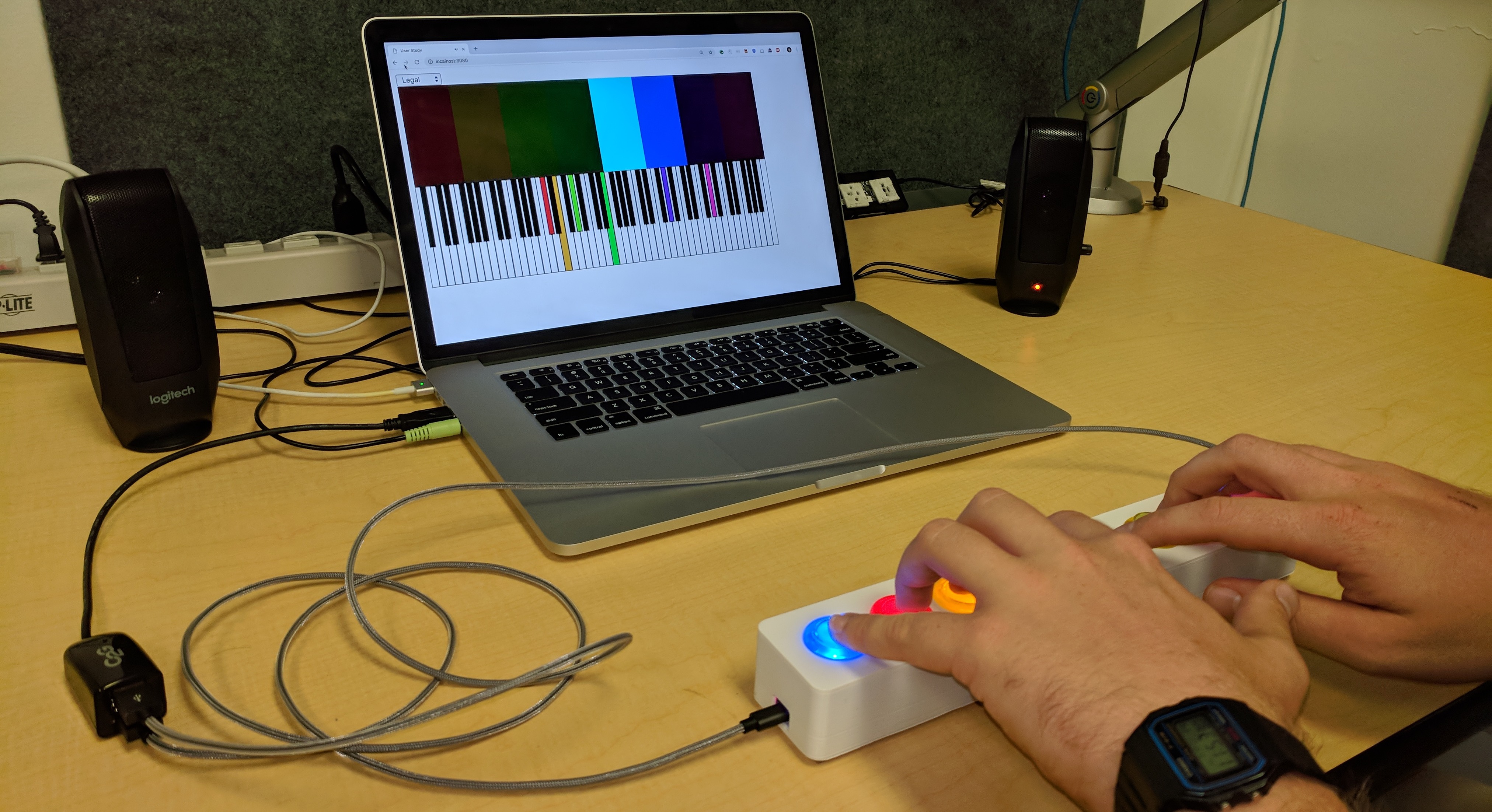}
    \caption{A user engages with the Piano Genie web interface during our user study.}
    \label{fig:userstudy}
\end{figure}

\begin{table}
\centering
\caption{Results for a small ($n=8$) user study for Piano Genie. Partipicants were given up to three minutes to improvise on three possible mappings between eight buttons and an $88$-key piano: 1)~(\emph{G-maj}) the eight buttons are mapped to a G-major scale, 2~(\emph{language model}) all buttons trigger our baseline language model, 3~(\emph{Piano Genie}) our proposed method. (\emph{Time}) is the average amount of time in seconds that users improvised with a mapping. (\emph{Per.}, \emph{Mus.}, \emph{Con.}) are the respective averages users expressed for enjoyment of performance experience, enjoyment of music, and level of control.}
\begin{tabular}{lcccc}
\toprule
Mapping & Time ($s$) & Per. & Mus. & Con. \\
\midrule
G-maj scale & $92.6$      & $3.125$ & $3.250$ & $4.625$ \\
Language model & $127.9$  & $3.750$ & $3.125$ & $1.750$ \\
Piano Genie & $144.1$     & $4.375$ & $3.125$ & $3.125$\\
\bottomrule
\end{tabular}
\label{tab:userstudy}
\end{table}

While our above analysis is useful as a sanity check, 
it offers limited intuition about how Piano Genie behaves in the hands of users. 
Accordingly, we designed a user study to compare three mappings between eight buttons and a piano: 
\begin{enumerate}[label=\arabic*.]
    \item (\emph{G-maj}) the eight buttons are deterministically mapped to a G major scale
    \item (\emph{language model}) pressing any button triggers a prediction by our baseline musical language model
    \item (\emph{Piano Genie}) our IQAE model with contour regularization
\end{enumerate}
Our reason for including the G-maj baseline is to gauge the performance experience of a mapping where a user has deterministic control but has to depend upon their own musical knowledge to produce patterns of interest. 
Our motivation for comparing Piano Genie to the language model is to determine how the performance experience is altered by including melodic control control in addition to rhythm.

Eight participants were given up to three minutes to improvise with each mapping (Figure~\ref{fig:userstudy}). 
The length of time they spent on each mapping was recorded as an implicit feedback signal. 
Participants reported a wide range of experience with piano performance: three had ``no experience'', half had ``some experience'', and one had ``substantial experience''. 
After each mapping, participants were asked to what extent they agreed with the following statements:
\begin{enumerate}[label=\Alph*.]
    \item ``I enjoyed the experience of performing this instrument''
    \item ``I enjoyed the music that was produced while I played''
    \item ``I was able to control the music that was produced''
\end{enumerate}
This survey was conducted on a five-level Likert scale~\cite{likert1932technique}
and we convert the responses to a $1$-$5$ numerical scale in order to compare averages (Table~\ref{tab:userstudy}).

When asked about their enjoyment of the performance experience, 
all eight users preferred Piano Genie to G-maj, 
while seven preferred Piano Genie to the language model. 
Five out of eight users enjoyed the music produced by Piano Genie more than that produced by G-maj. 
As expected, 
no participants said that Piano Genie gave them more control than the G-maj scale. 
However, 
all eight said that Piano Genie gave them more control than the language model. 

Though our user study was too limited in scope to make meaningful statistical claims, 
informally speaking the participants were quite enthusiastic about Piano Genie in comparison to the other mappings. 
One participant said ``there were some times when [Piano Genie] felt like it was reading my mind''.
Another participant said ``how you can cover the entire keyboard with only $8$ buttons is pretty cool.'' 
One mentioned that the generative component helped them overcome stage fright; they could blame Piano Genie for perceived errors and take credit for perceived successes. 
Several participants cited their inability to produce the same notes when playing the same button pattern as a potential drawback; enabling these patterns of repetition is a promising avenue for future work.
The participants with less piano experience said they would have liked some more instruction about types of gestures to perform. 

\section{Web demo details}

We built a web demo (polished demo: \url{https://piano-genie.glitch.me}; earlier version with all models from this paper: \url{https://bit.ly/2FaMeI4}) for Piano Genie to allow us to both improvise with our models and conduct our user study. 
Our web demo uses \textsc{TensorFlow.js}\footnote{\coollink{js.tensorflow.org}} to run our separately-trained neural networks in the browser in real-time. 
When a user presses a button, we pass this value into our trained decoder and run a forward pass producing a vector of $88$ logits representing the piano keys. 
We divide the logits by a \emph{temperature} parameter before normalizing them to a probability distribution with a \emph{softmax}. 
If the temperature is $0$, sampling from this distribution is equivalent to \emph{argmax}. 
Informally, we found a temperature of $0.25$ to yield a satisfying experience.

For the models that use the $\Delta{}T$ features, we have to wait until the user presses a key to run a forward pass of the neural network. 
For the models that do not use these features, we can run the computation for all $8$ possible buttons in advance. 
This allows us to both reduce the latency and display a helpful visualization of the possible model outputs contingent upon the user pressing any of the buttons (only available in our earlier demo \url{https://bit.ly/2FaMeI4}).

To build an interface for Piano Genie that would be more inviting than a computer keyboard, 
we 3D-printed enclosures for eight arcade buttons which communicate with the computer via USB (Figure~\ref{fig:interface}).\footnote{\coollink{learn.adafruit.com/arcade-button-control-box/overview}} 
Due to technical limitations of our USB microcontroller, we ended up building two boxes with four buttons instead of one with eight. 
This resulted in multiple unintended but interesting control modalities.
Several users rearranged the boxes from a straight line to different 2D configurations. 
Another user---a flutist---picked up the controllers and held them to their mouth. 
A pair of users each took a box and improvised a duet.

\section{Conclusion}

We have proposed Piano Genie, an intelligent controller which grants non-musicians a degree of piano improvisation literacy. 
Piano Genie has an immediacy not shared by other work in this space; 
sound is produced the moment a player interacts with our system rather than requiring laborious configuration. 
Additionally, the player is kept in the improvisational loop as they respond to the generative procedure in real-time. 
We believe that the autoencoder framework is a promising approach for learning mappings between complex interfaces and simpler ones, and hope that this work encourages future investigation of this space.

\begin{acks}
Special thanks to Monica Dinculescu for developing the polished web demo for Piano Genie. 
Thanks to 
Adam Roberts, 
Anna Huang, 
Ashish Vaswani, 
Ben Poole, 
Colin Raffel, 
Curtis Hawthorne, 
Doug Eck, 
Jesse Engel, 
Jon Gillick, 
Yingtao Tian 
and others at Google AI for helpful discussions and feedback throughout this work. 
We would also like to thank Julian McAuley, Stephen Merity, and Tejaswinee Kelkar for helpful conversations. 
Finally, we would like to thank all participants of our user study.
\end{acks}

\bibliographystyle{ACM-Reference-Format}
\bibliography{pianogenie}


\begin{thebibliography}{28}


\ifx \showCODEN    \undefined \def \showCODEN     #1{\unskip}     \fi
\ifx \showDOI      \undefined \def \showDOI       #1{#1}\fi
\ifx \showISBNx    \undefined \def \showISBNx     #1{\unskip}     \fi
\ifx \showISBNxiii \undefined \def \showISBNxiii  #1{\unskip}     \fi
\ifx \showISSN     \undefined \def \showISSN      #1{\unskip}     \fi
\ifx \showLCCN     \undefined \def \showLCCN      #1{\unskip}     \fi
\ifx \shownote     \undefined \def \shownote      #1{#1}          \fi
\ifx \showarticletitle \undefined \def \showarticletitle #1{#1}   \fi
\ifx \showURL      \undefined \def \showURL       {\relax}        \fi
\providecommand\bibfield[2]{#2}
\providecommand\bibinfo[2]{#2}
\providecommand\natexlab[1]{#1}
\providecommand\showeprint[2][]{arXiv:#2}

\bibitem[\protect\citeauthoryear{??}{pia}{2018}]%
        {pianoe}
 \bibinfo{year}{2018}\natexlab{}.
\newblock \bibinfo{title}{Piano-e-Competition}.
\newblock \bibinfo{howpublished}{\url{http://www.piano-e-competition.com/}}.
\newblock


\bibitem[\protect\citeauthoryear{Ball{\'e}, Laparra, and Simoncelli}{Ball{\'e}
  et~al\mbox{.}}{2017}]%
        {balle2016end}
\bibfield{author}{\bibinfo{person}{Johannes Ball{\'e}}, \bibinfo{person}{Valero
  Laparra}, {and} \bibinfo{person}{Eero~P Simoncelli}.}
  \bibinfo{year}{2017}\natexlab{}.
\newblock \showarticletitle{End-to-end optimized image compression}. In
  \bibinfo{booktitle}{\emph{ICLR}}.
\newblock


\bibitem[\protect\citeauthoryear{Bengio, L{\'e}onard, and Courville}{Bengio
  et~al\mbox{.}}{2013}]%
        {bengio2013estimating}
\bibfield{author}{\bibinfo{person}{Yoshua Bengio}, \bibinfo{person}{Nicholas
  L{\'e}onard}, {and} \bibinfo{person}{Aaron Courville}.}
  \bibinfo{year}{2013}\natexlab{}.
\newblock \showarticletitle{Estimating or propagating gradients through
  stochastic neurons for conditional computation}.
\newblock \bibinfo{journal}{\emph{arXiv:1308.3432}} (\bibinfo{year}{2013}).
\newblock


\bibitem[\protect\citeauthoryear{Bevilacqua, M{\"u}ller, and
  Schnell}{Bevilacqua et~al\mbox{.}}{2005}]%
        {bevilacqua2005mnm}
\bibfield{author}{\bibinfo{person}{Fr{\'e}d{\'e}ric Bevilacqua},
  \bibinfo{person}{R{\'e}my M{\"u}ller}, {and} \bibinfo{person}{Norbert
  Schnell}.} \bibinfo{year}{2005}\natexlab{}.
\newblock \showarticletitle{{MnM}: a {M}ax/{MSP} mapping toolbox}. In
  \bibinfo{booktitle}{\emph{NIME}}.
\newblock


\bibitem[\protect\citeauthoryear{Dowling}{Dowling}{1978}]%
        {dowling1978scale}
\bibfield{author}{\bibinfo{person}{W~Jay Dowling}.}
  \bibinfo{year}{1978}\natexlab{}.
\newblock \showarticletitle{Scale and contour: Two components of a theory of
  memory for melodies.}
\newblock \bibinfo{journal}{\emph{Psychological review}}
  (\bibinfo{year}{1978}).
\newblock


\bibitem[\protect\citeauthoryear{Eitan, Schupak, Gotler, and Marks}{Eitan
  et~al\mbox{.}}{2014}]%
        {eitan2014lower}
\bibfield{author}{\bibinfo{person}{Zohar Eitan}, \bibinfo{person}{Asi Schupak},
  \bibinfo{person}{Alex Gotler}, {and} \bibinfo{person}{Lawrence~E Marks}.}
  \bibinfo{year}{2014}\natexlab{}.
\newblock \showarticletitle{Lower pitch is larger, yet falling pitches shrink}.
\newblock \bibinfo{journal}{\emph{Experimental psychology}}
  (\bibinfo{year}{2014}).
\newblock


\bibitem[\protect\citeauthoryear{Fiebrink, Trueman, and Cook}{Fiebrink
  et~al\mbox{.}}{2009}]%
        {fiebrink2009meta}
\bibfield{author}{\bibinfo{person}{Rebecca Fiebrink}, \bibinfo{person}{Dan
  Trueman}, {and} \bibinfo{person}{Perry~R Cook}.}
  \bibinfo{year}{2009}\natexlab{}.
\newblock \showarticletitle{A Meta-Instrument for Interactive, On-the-Fly
  Machine Learning}. In \bibinfo{booktitle}{\emph{NIME}}.
\newblock


\bibitem[\protect\citeauthoryear{Gillian and Knapp}{Gillian and Knapp}{2011}]%
        {gillian2011machine}
\bibfield{author}{\bibinfo{person}{Nicholas Gillian} {and}
  \bibinfo{person}{Benjamin Knapp}.} \bibinfo{year}{2011}\natexlab{}.
\newblock \showarticletitle{A Machine Learning Toolbox For Musician Computer
  Interaction}. In \bibinfo{booktitle}{\emph{NIME}}.
\newblock


\bibitem[\protect\citeauthoryear{God{\o}y, Haga, and Jensenius}{God{\o}y
  et~al\mbox{.}}{2005}]%
        {godoy2005playing}
\bibfield{author}{\bibinfo{person}{Rolf~Inge God{\o}y}, \bibinfo{person}{Egil
  Haga}, {and} \bibinfo{person}{Alexander~Refsum Jensenius}.}
  \bibinfo{year}{2005}\natexlab{}.
\newblock \showarticletitle{Playing ``air instruments'': mimicry of
  sound-producing gestures by novices and experts}. In
  \bibinfo{booktitle}{\emph{International Gesture Workshop}}.
\newblock


\bibitem[\protect\citeauthoryear{God{\o}y and Jensenius}{God{\o}y and
  Jensenius}{2009}]%
        {godoy2009body}
\bibfield{author}{\bibinfo{person}{Rolf~Inge God{\o}y} {and}
  \bibinfo{person}{Alexander~Refsum Jensenius}.}
  \bibinfo{year}{2009}\natexlab{}.
\newblock \showarticletitle{Body movement in music information retrieval}.
\newblock  (\bibinfo{year}{2009}).
\newblock


\bibitem[\protect\citeauthoryear{Harmonix}{Harmonix}{1998}]%
        {theaxe}
\bibfield{author}{\bibinfo{person}{Harmonix}.} \bibinfo{year}{1998}\natexlab{}.
\newblock \bibinfo{title}{\emph{The Axe: Titans of Classic Rock}}.
\newblock \bibinfo{howpublished}{Game}.
\newblock


\bibitem[\protect\citeauthoryear{Harmonix}{Harmonix}{2005}]%
        {guitarhero}
\bibfield{author}{\bibinfo{person}{Harmonix}.} \bibinfo{year}{2005}\natexlab{}.
\newblock \bibinfo{title}{\emph{Guitar Hero}}.
\newblock \bibinfo{howpublished}{Game}.
\newblock


\bibitem[\protect\citeauthoryear{Hochreiter and Schmidhuber}{Hochreiter and
  Schmidhuber}{1997}]%
        {hochreiter1997long}
\bibfield{author}{\bibinfo{person}{Sepp Hochreiter} {and}
  \bibinfo{person}{J{\"u}rgen Schmidhuber}.} \bibinfo{year}{1997}\natexlab{}.
\newblock \showarticletitle{Long short-term memory}.
\newblock \bibinfo{journal}{\emph{Neural computation}} (\bibinfo{year}{1997}).
\newblock


\bibitem[\protect\citeauthoryear{Huron}{Huron}{1996}]%
        {huron1996melodic}
\bibfield{author}{\bibinfo{person}{David Huron}.}
  \bibinfo{year}{1996}\natexlab{}.
\newblock \showarticletitle{The melodic arch in Western folksongs}.
\newblock \bibinfo{journal}{\emph{Computing in Musicology}}
  (\bibinfo{year}{1996}).
\newblock


\bibitem[\protect\citeauthoryear{Kelkar and Jensenius}{Kelkar and
  Jensenius}{2017}]%
        {kelkar2017representation}
\bibfield{author}{\bibinfo{person}{Tejaswinee Kelkar} {and}
  \bibinfo{person}{Alexander~Refsum Jensenius}.}
  \bibinfo{year}{2017}\natexlab{}.
\newblock \showarticletitle{Representation Strategies in Two-handed Melodic
  Sound-Tracing}. In \bibinfo{booktitle}{\emph{International Conference on
  Movement Computing}}.
\newblock


\bibitem[\protect\citeauthoryear{Kelkar, Roy, and Jensenius}{Kelkar
  et~al\mbox{.}}{2018}]%
        {kelkar2018evaluating}
\bibfield{author}{\bibinfo{person}{Tejaswinee Kelkar}, \bibinfo{person}{Udit
  Roy}, {and} \bibinfo{person}{Alexander~Refsum Jensenius}.}
  \bibinfo{year}{2018}\natexlab{}.
\newblock \showarticletitle{Evaluating a collection of sound-tracing data of
  melodic phrases}. In \bibinfo{booktitle}{\emph{ISMIR}}.
\newblock


\bibitem[\protect\citeauthoryear{Kitahara, Giraldo, and Ram{\'\i}rez}{Kitahara
  et~al\mbox{.}}{2017}]%
        {kitahara2017jamsketch}
\bibfield{author}{\bibinfo{person}{Tetsuro Kitahara}, \bibinfo{person}{Sergio~I
  Giraldo}, {and} \bibinfo{person}{Rafael Ram{\'\i}rez}.}
  \bibinfo{year}{2017}\natexlab{}.
\newblock \showarticletitle{JamSketch: a drawing-based real-time evolutionary
  improvisation support system}. In \bibinfo{booktitle}{\emph{NIME}}.
\newblock


\bibitem[\protect\citeauthoryear{Lartilot}{Lartilot}{2018}]%
        {olivier2018soundtracer}
\bibfield{author}{\bibinfo{person}{Olivier Lartilot}.}
  \bibinfo{year}{2018}\natexlab{}.
\newblock \bibinfo{title}{\emph{Sound Tracer}}.
\newblock
\newblock


\bibitem[\protect\citeauthoryear{Lee, Freed, and Wessel}{Lee
  et~al\mbox{.}}{1992}]%
        {lee1992neural}
\bibfield{author}{\bibinfo{person}{Michael Lee}, \bibinfo{person}{Adrian
  Freed}, {and} \bibinfo{person}{David Wessel}.}
  \bibinfo{year}{1992}\natexlab{}.
\newblock \showarticletitle{Neural networks for simultaneous classification and
  parameter estimation in musical instrument control}. In
  \bibinfo{booktitle}{\emph{Adaptive and Learning Systems}}.
\newblock


\bibitem[\protect\citeauthoryear{Likert}{Likert}{1932}]%
        {likert1932technique}
\bibfield{author}{\bibinfo{person}{Rensis Likert}.}
  \bibinfo{year}{1932}\natexlab{}.
\newblock \showarticletitle{A technique for the measurement of attitudes}.
\newblock \bibinfo{journal}{\emph{Archives of psychology}}
  (\bibinfo{year}{1932}).
\newblock


\bibitem[\protect\citeauthoryear{Nymoen, Caramiaux, Kozak, and Torresen}{Nymoen
  et~al\mbox{.}}{2011}]%
        {nymoen2011analyzing}
\bibfield{author}{\bibinfo{person}{Kristian Nymoen}, \bibinfo{person}{Baptiste
  Caramiaux}, \bibinfo{person}{Mariusz Kozak}, {and} \bibinfo{person}{Jim
  Torresen}.} \bibinfo{year}{2011}\natexlab{}.
\newblock \showarticletitle{Analyzing sound tracings: a multimodal approach to
  music information retrieval}. In \bibinfo{booktitle}{\emph{ACM Workshop on
  Music Information Retrieval with User-centered and Multimodal Strategies}}.
\newblock


\bibitem[\protect\citeauthoryear{Parsons}{Parsons}{1975}]%
        {parsons1975directory}
\bibfield{author}{\bibinfo{person}{Denys Parsons}.}
  \bibinfo{year}{1975}\natexlab{}.
\newblock \bibinfo{booktitle}{\emph{The directory of tunes and musical
  themes}}.
\newblock


\bibitem[\protect\citeauthoryear{Roy, Kelkar, and Indurkhya}{Roy
  et~al\mbox{.}}{2014}]%
        {roy2014trap}
\bibfield{author}{\bibinfo{person}{Udit Roy}, \bibinfo{person}{Tejaswinee
  Kelkar}, {and} \bibinfo{person}{Bipin Indurkhya}.}
  \bibinfo{year}{2014}\natexlab{}.
\newblock \showarticletitle{TrAP: An Interactive System to Generate Valid Raga
  Phrases from Sound-Tracings.}. In \bibinfo{booktitle}{\emph{NIME}}.
\newblock


\bibitem[\protect\citeauthoryear{Scurto and Fiebrink}{Scurto and
  Fiebrink}{2016}]%
        {scurto2016grab}
\bibfield{author}{\bibinfo{person}{Hugo Scurto} {and} \bibinfo{person}{Rebecca
  Fiebrink}.} \bibinfo{year}{2016}\natexlab{}.
\newblock \showarticletitle{Grab-and-play mapping: Creative machine learning
  approaches for musical inclusion and exploration}. In
  \bibinfo{booktitle}{\emph{ICMC}}.
\newblock


\bibitem[\protect\citeauthoryear{Simon and Oore}{Simon and Oore}{2017}]%
        {performancernn2017}
\bibfield{author}{\bibinfo{person}{Ian Simon} {and} \bibinfo{person}{Sageev
  Oore}.} \bibinfo{year}{2017}\natexlab{}.
\newblock \bibinfo{title}{Performance RNN: Generating music with expressive
  timing and dynamics}.
\newblock
  \bibinfo{howpublished}{\url{https://magenta.tensorflow.org/performance-rnn}}.
\newblock


\bibitem[\protect\citeauthoryear{Theis, Shi, Cunningham, and Husz{\'a}r}{Theis
  et~al\mbox{.}}{2017}]%
        {theis2017lossy}
\bibfield{author}{\bibinfo{person}{Lucas Theis}, \bibinfo{person}{Wenzhe Shi},
  \bibinfo{person}{Andrew Cunningham}, {and} \bibinfo{person}{Ferenc
  Husz{\'a}r}.} \bibinfo{year}{2017}\natexlab{}.
\newblock \showarticletitle{Lossy image compression with compressive
  autoencoders}. In \bibinfo{booktitle}{\emph{ICLR}}.
\newblock


\bibitem[\protect\citeauthoryear{Trehub, Bull, and Thorpe}{Trehub
  et~al\mbox{.}}{1984}]%
        {trehub1984infants}
\bibfield{author}{\bibinfo{person}{Sandra~E Trehub}, \bibinfo{person}{Dale
  Bull}, {and} \bibinfo{person}{Leigh~A Thorpe}.}
  \bibinfo{year}{1984}\natexlab{}.
\newblock \showarticletitle{Infants' perception of melodies: The role of
  melodic contour}.
\newblock \bibinfo{journal}{\emph{Child development}} (\bibinfo{year}{1984}).
\newblock


\bibitem[\protect\citeauthoryear{van~den Oord, Vinyals, et~al\mbox{.}}{van~den
  Oord et~al\mbox{.}}{2017}]%
        {van2017neural}
\bibfield{author}{\bibinfo{person}{Aaron van~den Oord}, \bibinfo{person}{Oriol
  Vinyals}, {et~al\mbox{.}}} \bibinfo{year}{2017}\natexlab{}.
\newblock \showarticletitle{Neural discrete representation learning}. In
  \bibinfo{booktitle}{\emph{NIPS}}.
\newblock


\end{thebibliography}

\end{document}